\documentclass[lettersize,journal]{IEEEtran}
\usepackage{amsmath,amsfonts}
\usepackage{algorithmic}
\usepackage{algorithm}
\usepackage{array}
\usepackage{authblk}
\usepackage[caption=false,font=normalsize,labelfont=sf,textfont=sf]{subfig}
\usepackage{textcomp}
\usepackage{stfloats}
\usepackage{url}
\usepackage{verbatim}
\usepackage{graphicx}
\usepackage{multirow}
\usepackage[
backend=biber,
style=numeric,
sorting=ynt
]{biblatex}
\usepackage[nolist,nohyperlinks]{acronym}
\newacro{JAAD}{Joint Attention for Autonomous Driving}
\newacro{PIE}{Pedestrian Intention Estimation}
\newacro{KG}{Knowledge Graph}
\newacro{KGE}{Knowledge Graph Embedding}
\newacro{OccluRoads}{Occlusion-Rich Road Scenes with Pedestrians}
\newacro{ViT}{Vision Transformer}
\newacro{CNN}{Convolutional neural network}

\title{Prediction of Occluded Pedestrians in Road Scenes using Human-like Reasoning: Insights from the OccluRoads Dataset}

\author[1]{A. N. Melo}
\author[1]{S. Martín}
\author[1]{C. Salinas}
\author[1]{M. A. Sotelo}
\affil[1]{Department of Computer Engineering, Universidad de Alcalá, Madrid, Spain \\
[nataly.melo, sergio.martin, carlota.salinasmaldo, miguel.sotelo]@uah.es}
\begin{document}

\maketitle

\begin{abstract}

Pedestrian detection is a critical task in autonomous driving, aimed at enhancing safety and reducing risks on the road. Over recent years, significant advancements have been made in improving detection performance. However, these achievements still fall short of human perception, particularly in cases involving occluded pedestrians, especially entirely invisible ones. In this work, we present the \ac{OccluRoads} dataset, which features a diverse collection of road scenes with partially and fully occluded pedestrians in both real and virtual environments. All scenes are meticulously labeled and enriched with contextual information that encapsulates human perception in such scenarios. Using this dataset, we developed a pipeline to predict the presence of occluded pedestrians, leveraging \ac{KG}, \ac{KGE}, and a Bayesian inference process. Our approach achieves a F1 score of 0.91, representing an improvement of up to 42\% compared to traditional machine learning models.

\end{abstract}


\section{Introduction}
In recent years, self-driving cars have made rapid advancements in both driving complexity and safety. However, recent reports on pedestrian road accidents reveal significant challenges that pedestrians face globally. According to the European Road Safety Observatory (ERSO), pedestrians represent around 20\% of all road fatalities, with higher fatality rates in urban areas, where interactions with vehicles are more frequent \cite{erso2022}. Similarly, the World Health Organization's Global Status Report on Road Safety highlights that pedestrian fatalities make up a substantial portion of global road deaths, especially in low and middle income countries, where pedestrians often account for over 21\% of road traffic fatalities \cite{who}.

Given these statistics, predicting the behavior of road users is crucial for risk management and improving driving safety. Although significant progress has been made in pedestrian detection, performance remains limited when pedestrians are occluded. Occlusions affected by factors such as lighting, weather, environment, and road scene surroundings often result in missed detections and reduced detection accuracy. The occlusion challenge in pedestrian 
\begin{figure}[ht]
\centering
\includegraphics[scale=0.35]{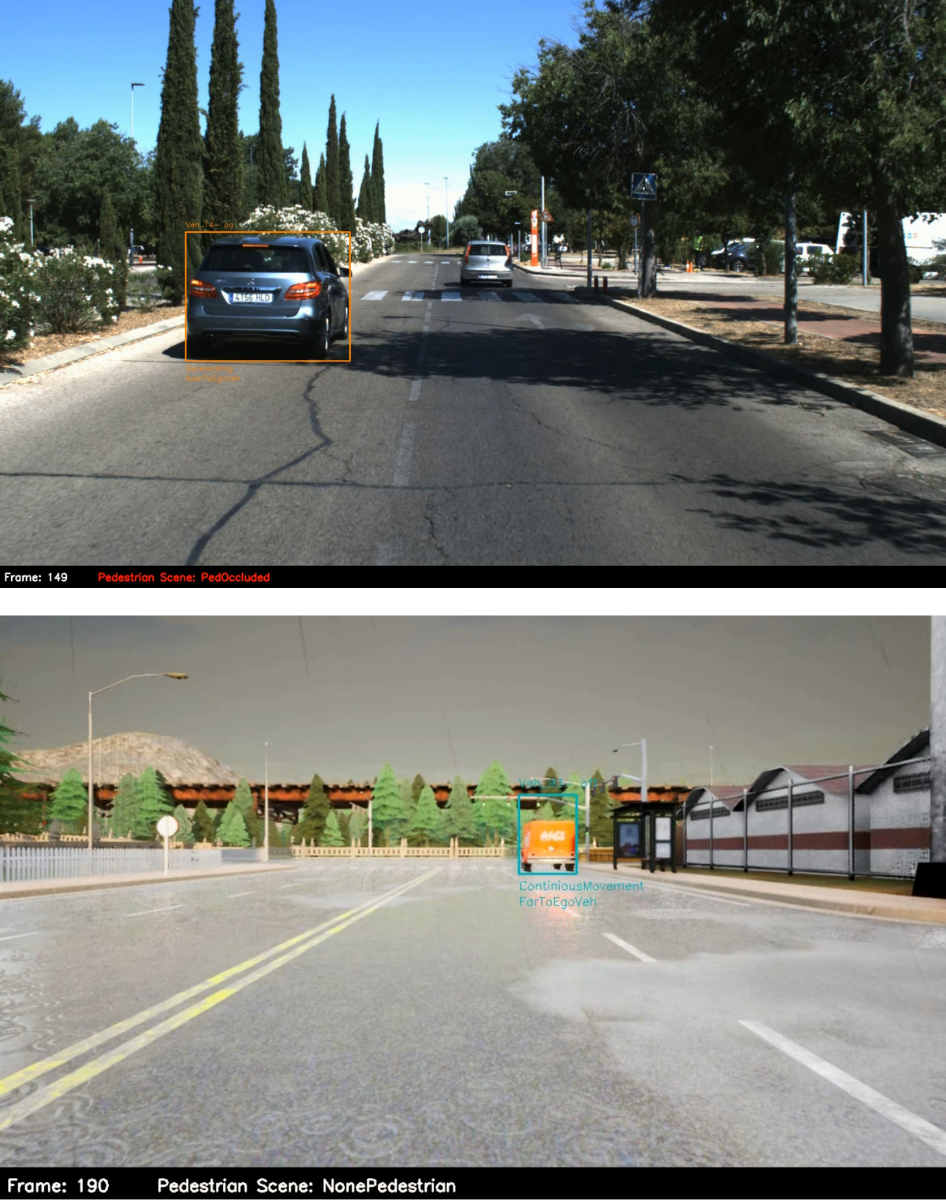}
\caption{OccluRoads dataset examples}
\label{fig:examples-dataset}
\end{figure}
detection primarily involves two types: intra-class occlusions (between pedestrians) 
and occlusions caused by other objects in the scene \cite{occlusion-review}. Full occlusions, however, are seldom addressed in current research, as they present additional complexities and limited strategic approaches.

Furthermore, while existing pedestrian detection benchmarks provide annotations for partial occlusions, they vary significantly in defining the occurrence and severity of occlusion. Notably, no dataset focuses exclusively on road scenarios with occluded pedestrians, especially fully occluded ones.

In this study, we aim to address the need to predict the presence of occluded pedestrians in road scenes by leveraging a knowledge-based approach that incorporates cues from human perception, labeled in our own dataset, named \ac{OccluRoads}. Figure \ref{fig:examples-dataset}
shows an example from the dataset in both real and virtual environments. The contributions of this work are threefold: (1) the creation of a robust road scene dataset featuring occluded pedestrians, (2) the introduction of an occluded pedestrian predictor based on a knowledge-driven approach, and (3) an evaluation of the proposed predictor on the created dataset.

The remainder of the article is organized as follows: Section II discusses related work, Section III details the dataset and methodology, Section IV describes the experiments and implementation, Section V presents results and discussion, and Section VI concludes the study with insights for future research.

\section{Related work}

\subsection{Pedestrian occlusion datasets}
In the context of autonomous driving, various datasets provide pedestrian annotations, primarily focused on pedestrian detection and crossing behavior. However, there is no dataset specifically dedicated to capture pedestrian occlusion from a vehicle’s perspective, especially with an emphasis on pedestrians who are completely hidden from view.

A select group of datasets includes some level of occlusion labeling for pedestrians, with the following being among the most prominent:

\begin{itemize}
    \item \ac{JAAD}\cite{Jaad} and \ac{PIE}\cite{pie}: These two datasets are widely used in pedestrian research and provide three occlusion flags: (1) none, (2) partial, and (3) full. Partial occlusion is defined when 25-75\% of the pedestrian is obscured, while full occlusion refers to less than 25
    \item KITTI\cite{kitti}:  The KITTI Vision Benchmark Suite, used extensively in autonomous driving research, includes occlusion levels categorized into four types: 0 (fully visible), 1 (partly occluded), 2 (largely occluded), and 3 (unknown level of occlusion).
    \item Caltech pedestrian\cite{Caltech}:  As one of the foundational datasets for pedestrian detection, it includes binary occlusion labels: 0 for not occluded and 1 for occluded. This dataset captures pedestrians who are never occluded, occasionally occluded, or consistently occluded across frames, with approximately 60\% of pedestrians being fully visible.
    \item CityPersons\cite{citypersons}: A subset of the Cityscapes dataset, CityPersons was recorded in highly populated urban environments. Less than 30\% of pedestrians are fully visible, and occlusion is represented by percentages, denoting levels of no occlusion, partial occlusion, and heavy occlusion.
\end{itemize}

Although these datasets capture pedestrian occlusion to varying degrees, they generally lack substantial representation of fully occluded pedestrians, especially in pedestrian crossing scenarios.

\subsection{Pedestrian occlusion prediction}
There has been extensive research focused on pedestrian detection, with significant improvements in performance, particularly for pedestrians with high visibility and minimal occlusion. However, detecting pedestrians who are partially or fully occluded remains a more challenging task, and some studies have specifically addressed this issue. In \cite{occlusion-review}, the authors highlight several approaches for detecting occluded pedestrians, such as utilizing context information, attentional mechanisms, body parts, multi-scale features, and feature fusion techniques.

Regarding body part-based approaches, a head-aware pedestrian detection network (HAPNet) was proposed in 2022, focusing on the structural relationship between the human body and the head. Using this association, HAPNet incorporates a scoring module and an augmented non-maximum suppression (NMS) algorithm to detect occluded pedestrians \cite{HAPnet}. In \cite{body-detection-semantic}, a method based on semantic relationships between body parts was used to detect partial occlusions of the body, although this approach was not applied in a road context. Additionally, some strategies approach the problem as a multi-class classification task, learning to identify body parts and relate them to occluded pedestrians, enabling detection from body parts alone \cite{body-fusion}.

Other research highlights the use of additional contextual features to improve the detection of occluded pedestrians. Many of these methods incorporate regional features to gather more information about the scene. For example, \cite{channel-features} represents pedestrians using a combination of channel features, where partial occlusion is addressed by constructing a hierarchical region reduction structure, followed by pedestrian detection using boosted decision trees. Another approach focuses on extracting more discriminative features of pedestrians and combining them with global context information derived from the visible part of the pedestrian \cite{global-fusion}.

Similarly, an approach has been proposed that combines semantic features of pedestrians with the merging of feature maps of different layers, together with a head detector, to detect pedestrians based on their locations and scales \cite{semantic-features}.

Overall, while the performance of the aforementioned approaches is promising and introduces novel techniques, they often do not account for fully occluded pedestrians. This limitation arises due to the lack of sufficient examples of fully occluded pedestrians in the commonly used datasets, which hinders further development in this area.

\section{Methodology}\label{AA}
 In this section, the methodology employed to build the \ac{OccluRoads} dataset is detailed. First the dataset collection and then the process of data labeling

\subsection{Dataset Collection}
Our dataset comprises over 99 video clips featuring road scenes both occluded and non-occluded pedestrians, capturing cases where occluded pedestrians become visible after several frames. Each scene lasting between 9 and 40 seconds. Of these, 40 videos were recorded in Spain using a Grasshopper3 camera mounted inside the car, positioned below the rearview mirror. The recordings were made at 25 FPS with a resolution of 1920x1200. 

The remaining videos were generated using the open-source autonomous driving simulator Carla \cite{carla}. Instead of performing pedestrian navigation through the simulator's deterministic or randomized control functions, a Virtual Reality framework was employed to integrate behaviors from real pedestrians, improving the authenticity of their interactions with vehicles \cite{CHIRA22}. The highly customizable blueprints available in CARLA library enabled the simulation of virtual cameras with the identical 1200x900 resolution at 25 FPS,  while also allowing the inclusion of various vehicle models and diverse weather conditions \cite{carla_sensor_blueprints}.

\subsection{Data Labeling}
\begin{figure}
\centering
\includegraphics[scale=0.60]{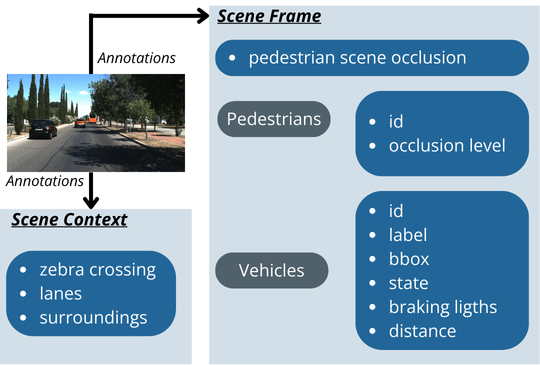}
\caption{Dataset Annotations}
\label{fig:annotations}
\end{figure}

\begin{figure*}
\centering
\includegraphics[width=0.95\textwidth]{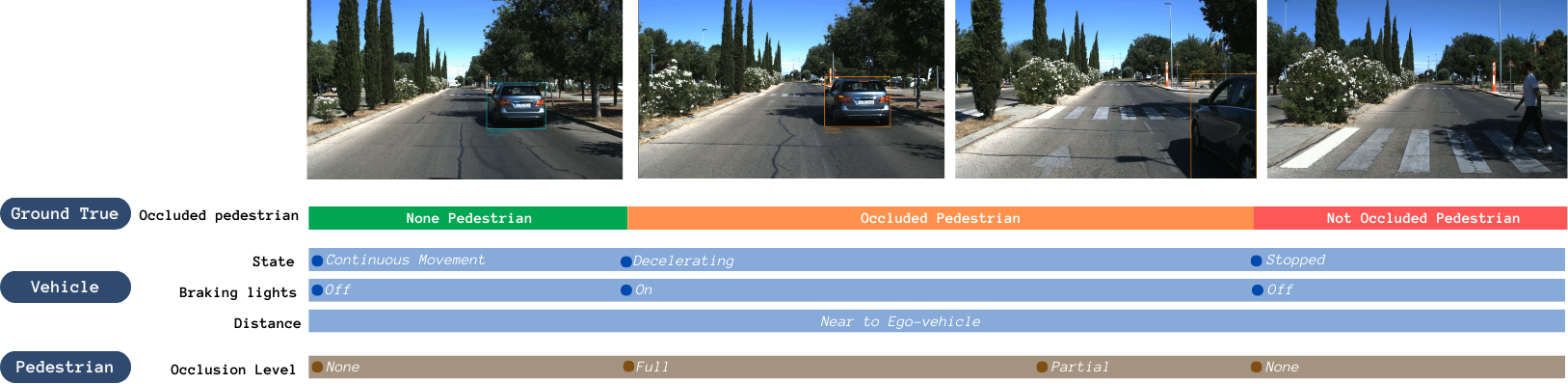}
\caption{Annotations to represent the contextual information of a road scene}
\label{fig:sequence-roadscene}
\end{figure*}

Focusing specifically on pedestrian occlusion within road scene scenarios, the dataset includes two main categories of labels: \textbf{Scene Context} and \textbf{Scene Frame}. The Scene Context category offers a general overview of each road scenario, while the Scene Frame category provides frame-by-frame information about occluded and non-occluded pedestrians, as well as the vehicles present in the scene. Figure \ref{fig:annotations} illustrates the main structure and labeling scheme of the dataset. Labels related to the scene context are provided once per road scene, while frame-specific features are annotated for each individual frame within the road scene and include data from pedestrians and vehicles. The annotations are stored in Extensible Markup Language (XML) format. Figure \ref{fig:sequence-roadscene} shows an example of all the annotations included in a road scenario.

The data dictionary, described in table \ref{tab-linguistic}, outlines the descriptions and possible values for each primary label. For both recorded and simulated videos, data labeling was conducted based on specific features, with this initial version relying on a combination of manual labeling, neural networks, and data from the Carla environment. The labeling process involved a primary labeling phase, followed by a review to ensure accuracy, as is described in the following lines:

\begin{table}[htbp]
\caption{Dataset data dictionary}
\begin{center}
\begin{tabular}{| p{1.9cm} | p{3.0cm} | p{2.3cm} |}
\hline
\textbf{Label}  & \textbf{Description} & \textbf{Values} \\
\hline
zebra\_crossing  & Presence of a zebra crossing in the scene & True, False \\
\hline
lanes  & Number of lanes in the scene & Number. Eg.2 \\
\hline
surroundings  & Characteristics of the surroundings & Vegetation, Clear \\
\hline
pedestrians\_scene  & Presence of an occluded pedestrian & NonePedestrian, PedestrianOccluded, PedestrianNotOccluded \\
\hline
state  & Describe the movement of a vehicle in the scene & ContinuousMovement, Stopped, Accelerating, Decelerating \\
\hline
braking\_ligths  & Describe the state of the braking lights of the vehicle & On, Off \\
\hline
distance  & Represents the estimated distance to the ego-vehicle & NearToEgoVeh, MiddleDisToEgoVeh, FarToEgoVeh \\
\hline
occlusion  & Level of pedestrian occlusion & None, Full, Partial \\
\hline
\end{tabular}
\label{tab-linguistic}
\end{center}
\end{table}

\begin{itemize}
\item \textbf{Scene Context Data:} Information such as zebra crossings, lane markings, and surrounding details were manually labeled for both recorded and simulated videos.
\item \textbf{Pedestrian scene occlusion:} To identify the presence of pedestrians in each frame, we utilized the YOLOv8 neural network \cite{YOLOV8}. If YOLOv8 detected a pedestrian, the pedestrian was labeled as “pedestrian not occluded.” If no pedestrian was detected, a manual verification was performed. If a pedestrian was present in the scene during verification, the label was updated to “occluded pedestrian”; otherwise, it was labeled as “none pedestrian”.
\item \textbf{Pedestrian Data:}  For cases where YOLOv8 detected a pedestrian, the occlusion level was automatically labeled as "None." However, when YOLOv8 did not detect a pedestrian but manual verification confirmed the presence of one in the road scene, occlusion details were manually annotated. If less than 25\% of the pedestrian body was visible, the occlusion level was labeled as “fully occluded”; otherwise, it was labeled as “partially occluded.” Each pedestrian was assigned a unique ID to enable consistent tracking across frames.
\item \textbf{Vehicle Data:} Each vehicle present in the scene for each frame was labeled with a unique ID and a position label indicating its relative location to the recording vehicle (e.g., "in front," "to the left," or "to the right").
A YOLO-based\cite{braking} neural network detected the vehicle’s braking lights and created bounding boxes around each vehicle. The braking lights status was automatically labeled based on this detection. For the recorded videos, the braking lights served as indicators for the vehicle’s state, which was cross-referenced by reviewing the video footage. In Carla environment, we relied on recorded vehicle speed data, using it to label the vehicle’s movement state.

The distance label was computed using triangle similarity, as represented by the equation:
\begin{equation}
 D = (W x F ) /P
\end{equation}
where W is the known width of the pedestrian, F is the camera’s focal length, and P is the pedestrian’s width in pixels.
\end{itemize}

\section{Experiments}
In this section we describe the experimental evaluation of the pedestrians occlusion detection.
\subsection{Model architecture} \label{sec:modelArchitecture}
\begin{figure}[ht]
\centering
\includegraphics[scale=0.55]{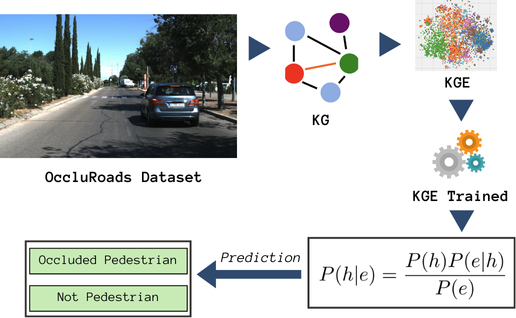}
\caption{Knowledge-based pipeline}
\label{fig:architecture}
\end{figure}

The approach used to predict the presence of an occluded pedestrian was based on knowledge and the pipeline presented in our previous work \cite{KGPedestrian}. The proposed knowledge-based  follows a pipeline (See Figure \ref{fig:architecture}) with three phases: 1) \ac{KG} generation, which creates a structured KG from features included in created dataset; 2) \ac{KGE} learning, which embeds the KG for further processing; and 3) Bayesian inference and prediction, which uses these embeddings to predict if there is a pedestrian occluded.

\subsubsection{KG Generation}
\begin{figure}[ht]
\includegraphics[scale=0.35]{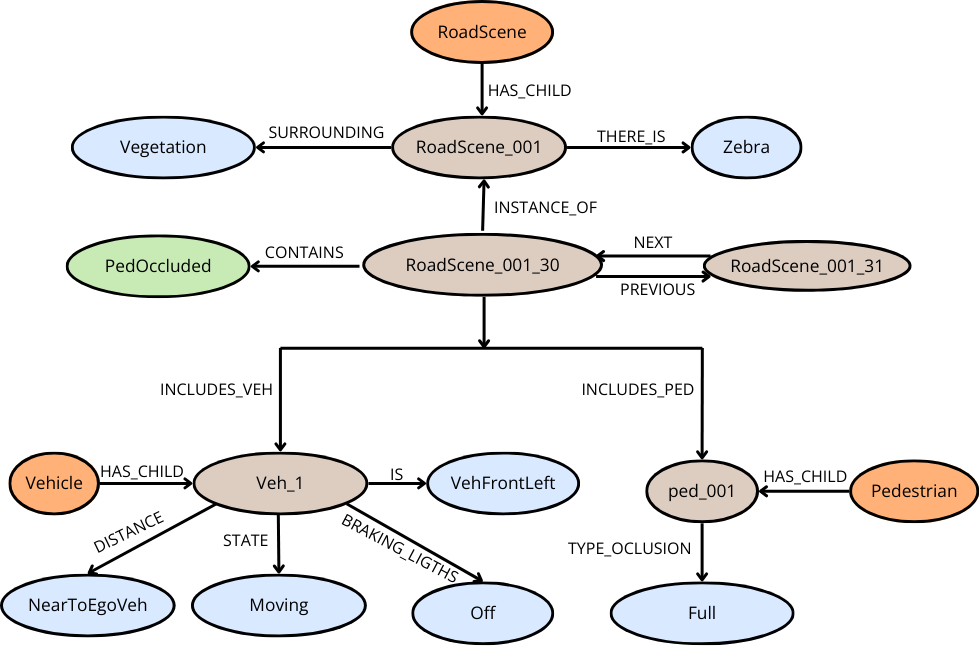}
\caption{Occluded Pedestrian Ontology}
\label{fig:ontology}
\end{figure}

In the step of generating a \ac{KG} to encapsulate data from the created dataset, an occluded pedestrian ontology was defined, as shown in figure \ref{fig:ontology}. The generated ontology includes three main entities: Pedestrian (representing all pedestrians), Vehicle (representing all vehicles), and RoadScene (representing each road scene or video). In the KG, each road scene is identified by a unique ID and is associated with contextual information, such as scene context, surroundings, and zebra crossing presence.

For each frame within a road scene, an entity called "RoadSceneFrame" is created, identified by an ID and frame number. Each "RoadSceneFrame" is linked to the previous and next frames, as well as to the pedestrian occlusion state (Pedestrian Occluded or Not Pedestrian) and information on pedestrians and vehicles relevant to that scene.

Within the ontology, each observed vehicle is linked to details such as its distance from the ego-vehicle, movement state, position relative to the ego-vehicle, and braking lights status. Similarly, each pedestrian entity is linked to the level of occlusion. This structure enables the KG to comprehensively represent occlusion dynamics and interactions in road scenes.

Based on these KG ontology and the data defined in \ac{OccluRoads}, the \ac{KG} is generated using the Ampligraph 2.0.0 library \cite{ampligraph}. The KG is formed in shape of triples and a group of triples represents the pedestrian state in a frame.

\subsubsection{KGE Learning}

During the second phase, Ampligraph 2.0.0 is utilized to develop a KGE model from the \ac{KG} constructed in the prior phase. Considering the nature of the KG's relations, the \textit{ComplEx} model is employed. Subsequently, this KGE model undergoes training, validation, and testing using Ampligraph's learning capabilities.

\subsubsection{Bayesian inference and Prediction}
Similar to the approach in \cite{KGPedestrian}, our method applies Bayesian inference to embeddings learned from Knowledge Graph Embedding (KGE) models. This enables inductive reasoning to predict the presence of occluded pedestrians who are not explicitly represented in the KG. We employ Bayes' rule, as shown in the following equation:
\begin{equation}
\label{eq_bayes}
    \begin{split}
        P(h|e)= \frac{P(h)P(e|h)}{P(e)}
    \end{split}
\end{equation} 
Here, the hypothesis \textit{h} represents the event or entity we aim to predict (i.e., the presence of an occluded pedestrian), while the evidence \textit{e} consists of contextual information about the road scene and related vehicles. Each component of the equation is computed by evaluating individual triples using the KGE evaluation method provided by Ampligraph. For this use case, we illustrate an example of occluded pedestrian prediction:
\begin{itemize}
\item \textbf{Hypothesis:} There is an occluded pedestrian in the road scene.
\item  \textbf{Evidence:} The road scene includes a zebra crossing, vegetation surrounds the road scene, a vehicle is positioned in the front-left of the road, close to the ego-vehicle, it is decelerating, and its braking lights are activated.
\end{itemize}

For this example, the computation of \(\ P(h)\) involves evaluating the triple \textless \textit{RoadScene}, contains, \textit{OccludedPed}\textgreater. The computation of \(\ P(e)\) considers the number of evidences extracted from the road scene, such as \textless \textit{RoadScene}, thereIs, \textit{ZebraCrossing}\textgreater,\textless \textit{RoadScene}, includes, \textit{VehDecelerating}\textgreater, and so on. Finally, \(\ P(e|h) \) is computed for each individual piece of evidence. With these values, \(\ P(h|e) \) can be calculated using Equation \ref{eq_bayes}, as all individual probabilities are derived from the KG using the learned embeddings.
\subsection{Implementation}
To implement our pipeline, as described in Section \ref{sec:modelArchitecture}, we used Python, TensorFlow, and the Ampligraph library. For training, we applied the ComplEx scoring model, along with the Adam optimizer and SelfAdversarialLoss. The training parameters included an embedding size of \(k = 150 \) , with 15 corruptions generated per training instance. Additionally, we set the learning rate to \(learningRate = 0.0005 \), the batch size to \(batchSize = 8000 \), used early stopping based on the mean reciprocal rank (MRR).
All the learnable methods were trained with a CPU AMD Ryzen 5 5600X 6-Core with a GPU NVIDIA GeForce RTX 3080.
\subsection{Experimental setup}
To test the knowledge-based approach on the \ac{OccluRoads} dataset, we designed an experimental setup to evaluate occluded pedestrian detection and assess the impact of training on each environment (real and virtual).  Consequently, we split the videos for training and testing as shown in Table \ref{tab:dataset-sampling}.

\begin{table}[htbp]
\caption{OccluRoads dataset sampling}
\begin{center}
\begin{tabular}{ p{2.0cm}  p{1.5cm} p{1.5cm} p{1.5cm} }
\textbf{Enviroment}  & \textbf{\# Videos} & \textbf{\# Train} & \textbf{\# Test} \\
\hline
Real  & 40 & 32 & 8 \\
\hline
Virtual  & 59 & 50 & 9 \\
\hline
\end{tabular}
\label{tab:dataset-sampling}
\end{center}
\end{table}
It is important to highlight that in our study, we predict the next 30th frame. To assess the pipeline’s performance, we used precision, recall, and F1-score metrics. Precision is defined as the ratio of correct positive predictions to total predicted positives. Recall represents the ratio of correct positive predictions to total positive instances, and the F1-score is the harmonic mean of precision and recall.

\section{Results and Discussions}
In this section an analysis of the created dataset is presented and also the results over the occluded pedestrian prediction, compared with different methods of validation and evaluated in a cross-enviroment context.

\subsection{Dataset analysis}
The \ac{OccluRoads} dataset consists of 70 videos with occluded pedestrians and 29 videos without any pedestrians, totaling 8,459 frames with occluded pedestrians, 9,735 frames with non-occluded pedestrians, and 21,520 frames without pedestrians. The dataset annotations include labels capturing occluded pedestrian ground truth as well as various contextual and behavioral features associated with each scenario.

\begin{figure}[ht]
\includegraphics[scale=0.28]{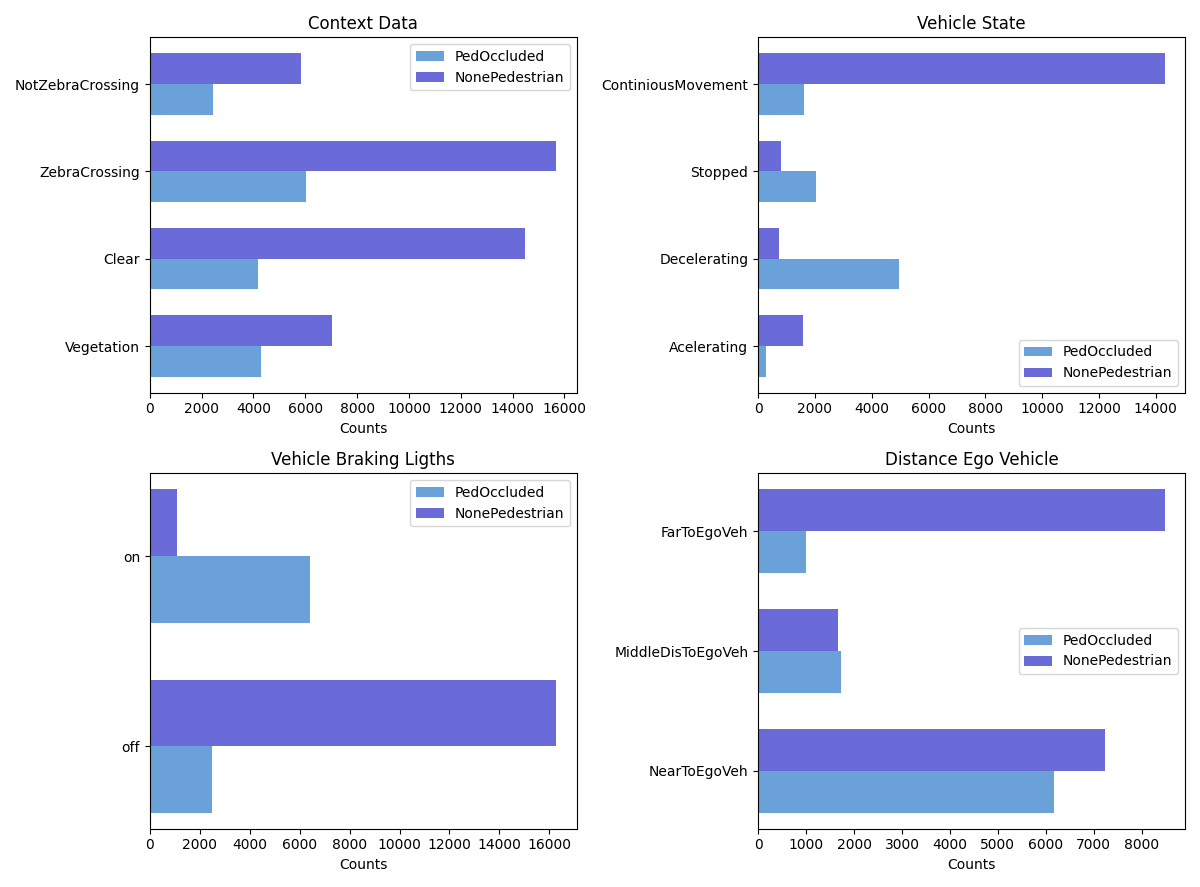}
\caption{OccluRoads data distribution}
\label{fig:data_distribution}
\end{figure}

As shown in Figure \ref{fig:data_distribution}, certain features are strongly associated with each pedestrian scene type. For scenarios without pedestrians, vehicles are typically in continuous motion with braking lights turned off. In contrast, scenes with occluded pedestrians generally involve vehicles decelerating, often with activated braking lights.

Regarding the distance between the vehicle and the ego-vehicle, a notable distinction is observed when vehicles are far from the ego-vehicle, as these instances are frequently associated with the absence of pedestrians.

In terms of contextual features, vegetation in the surroundings provides a clear visual distinction: scenes without vegetation tend to have fewer instances of fully occluded pedestrians, suggesting that open road scenes improve pedestrian visibility. On the other hand, the presence of zebra crossings shows mixed patterns while zebra crossings are more common in scenes without pedestrians, they are also present in cases with occluded pedestrians.

This analysis highlights the significance of vehicle behavior in predicting occluded pedestrians, particularly when pedestrians are entirely out of sight. Additionally, contextual features like zebra crossings and surroundings information are less informative on their own and may require additional contextual data to enhance predictive accuracy.

\subsection{Cross-environment validation prediction}
We evaluated the performance of the proposed approach on the \ac{OccluRoads} dataset, comparing results across three main scenarios: (1) training on real videos, (2) training on virtual videos, and (3) training on a combination of real and virtual videos. Each scenario was tested on both real and virtual test sets.

\begin{table}[h!]
\centering
\caption{Comparing the occlusion detection based on different training data}
\label{table:results}
\begin{tabular}{lcrl}
Train Data & F1  & Precision & Recall \\
\hline
Real & 0.86 & 0.81 & \textbf{0.92} \\
Virtual & \textbf{0.89} & 0.87 & 0.91 \\
Mixed & 0.77 & 0.63 &  \textbf{1} \\
\hline
\end{tabular}
\end{table}

As shown in Table \ref{table:results}, the model trained solely on data generated in  Carla environment achieves a higher F1 score when tested on both real and virtual environments. For testing on real data, the performance difference between models trained on real vs. virtual data is minimal, while testing on virtual data reveals a more significant difference, favoring the virtual-trained model. Overall, virtual data offers more consistent detection results and appears to contain features that better generalize the use case.

In contrast, the model trained on a mix of real and virtual data shows a significant performance decrease when tested on real data, but performs better in virtual environment.

\begin{table}[h!]
\centering
\begin{tabular}{lcrl}
Train Data & F1  & Precision & Recall \\
\hline
Real & 0.74 & 0.82 & 0.68 \\
Virtual & \textbf{0.94} & \textbf{0.94} & 0.94 \\
Mixed & 0.86 & 0.75 & \textbf{1} \\
\hline
\end{tabular}
\end{table}

Additionally, we trained the model using both real and virtual environments and then tested it across all cross-environment videos. As shown in table \ref{table:results-cross}, the model trained with virtual environment data performs better in cross-testing scenarios. This suggests that the supplementary data provided by the Carla environment, such as vehicle speed and braking status, enables more accurate labeling and consequently more reliable training data. These findings underscore the importance of deploying scenarios that incorporate additional data acquisition elements, which enhance the information available about road scenes and pedestrian occlusion.

\begin{table}[h!]
\centering
\caption{Cross-environment testing}
\label{table:results-cross}
\begin{tabular}{llcrl}
Test set & Train Data & F1  & Precision & Recall \\
\hline
Real & Virtual & 0.80 & 0.71 & 0.91 \\
Virtual & Real & 0.74 & 0.82 & 0.68 \\
\hline
\end{tabular}
\bigskip

\end{table}
On the following website [https://occluroads.s3.us-west-2.amazonaws.com/index.html], we provide access to the dataset collection, including both the videos and their annotations. Additionally, the website features examples of occluded pedestrian predictions using the OccluRoads dataset.

\subsection{Prediction methods validation}
To compare the impact of the approach based on knowledge with more traditional machine learning approaches, we trained a \ac{ViT} model \cite{ViT} which uses a Transformer architecture to process image data and includes a leverage self-attention mechanisms to capture relationships between different parts of an image, achieving impressive results on computer vision tasks such as image classification, object detection, and segmentation. Similarly, we trained a \ac{CNN} using a ResNet50 model, incorporating a fine-tuning process to predict the presence of occluded pedestrians based on image data.

\begin{table}[h!]
\centering
\caption{Comparing the occlusion detection with different methods}
\label{table:results-methods}
\begin{tabular}{lcrl}
Method & F1  & Precision & Recall \\
\hline
ViT & 0.49 & 0.48 & 0.49 \\
CNN & 0.51 & 0.55 & 0.48 \\
Ours & 0.91 & 0.89 & 0.93 \\
\hline
\end{tabular}
\end{table}

The results presented in Table \ref{table:results-methods} highlight the significant impact of incorporating contextual features in occluded pedestrian prediction compared to approaches that rely solely on road scene images. Our proposed method achieves up to 42\% improvement in F1 score, demonstrating the effectiveness of contextual information. This also underscores the advantages of a knowledge-based approach, which excels in scenarios where human perception is crucial. Specifically, in cases of full occlusion, the most accurate method for predicting the presence of occluded pedestrians involves leveraging available data alongside a comprehensive understanding of the road scene.

\section{Conclusions and Future work}

The dataset presented in this work represents a significant effort to advance research in the detection and prediction of occluded pedestrians. One of the key benefits of our dataset is its unique focus on road scenarios featuring occluded pedestrians, which distinguishes it from existing datasets. Additionally, it provides rich contextual labels that capture the entire road scenario. This initial version includes road scenes from both real and virtual environments. The importance of data quality is emphasized, as virtual reality frameworks like CARLA enable the generation of highly accurate data, with measurable and controllable interactions between all actors in the road scene.

This work also introduces a novel approach to predict the presence of occluded pedestrians using knowledge-driven methods, leveraging \ac{KG}, \ac{KGE}, and Bayesian inference. Cross-environment validation highlights the relevance of virtual environment simulations, demonstrating that our knowledge-based approach achieves a deeper understanding of the road scene. This enables accurate predictions of occluded pedestrians based on contextual information. Furthermore, the implemented pipeline improves prediction accuracy by up to 42\% compared to traditional methods that primarily rely on image-based features.

Future work will focus on expanding the dataset by incorporating more diverse road scenarios in both real and virtual environments. This will pave the way for establishing an occluded pedestrian prediction benchmark with curated sequences to challenge and engage the scientific community.

\section*{Acknowledgment}
This research has been funded by the HEIDI project of the European Commission under Grant Agreement: 101069538

\printbibliography

\end{document}